%% file: main.tex
\documentclass[10pt,twocolumn,letterpaper]{article}

\usepackage[pagenumbers]{iccv} 
\usepackage{pifont}
\usepackage{tabularx}
\usepackage{multirow}
\usepackage{graphicx}
\usepackage{caption}
\usepackage{array}
\usepackage{adjustbox} 
\usepackage{color, colortbl}
\definecolor{Gray}{rgb}{0.9,0.9,1}
\definecolor{LightGreen}{rgb}{0.6, 1, 0.6}
\definecolor{PastelBlue}{rgb}{0.68, 0.85, 0.9} 
\definecolor{PastelGreen}{rgb}{0.7, 1.0, 0.8} 
\definecolor{PastelPink}{rgb}{1.0, 0.85, 0.95} 
\definecolor{PastelPurple}{rgb}{0.93, 0.89, 1.0} 
\definecolor{PastelYellow}{rgb}{1.0, 1.0, 0.95} 
\definecolor{LighterPastelBlue}{rgb}{0.92, 0.96, 0.99} 
\definecolor{LighterPastelGreen}{rgb}{0.88, 1.0, 0.88} 
\definecolor{LighterPastelPink}{rgb}{1.0, 0.93, 0.95} 
\definecolor{LighterPastelPurple}{rgb}{0.96, 0.92, 1.0} 
\definecolor{LighterPastelYellow}{rgb}{1.0, 1.0, 0.9} 
\definecolor{LightOrange}{rgb}{1.0, 0.94, 0.86}

\definecolor{iccvblue}{rgb}{0.21,0.49,0.74}
\usepackage[pagebackref,breaklinks,colorlinks,allcolors=iccvblue]{hyperref}

\def\paperID{10506} 
\def\confName{ICCV}
\def\confYear{2025}

\title{Towards Scalable Modeling of Compressed Videos for Efficient Action Recognition}

\author{Shristi Das Biswas, Efstathia Soufleri, Arani Roy, Kaushik Roy\\
Purdue University\\
{\tt\small \{sdasbisw, roy173, kaushik\}@purdue.edu; esoufler@alumni.purdue.edu}
}
\normalsize
\begin{document}
\twocolumn[{
\renewcommand\twocolumn[1][]{#1}
\maketitle
\begin{center}
    \centering 
    \includegraphics[width=0.96\textwidth, height=0.37\textwidth]{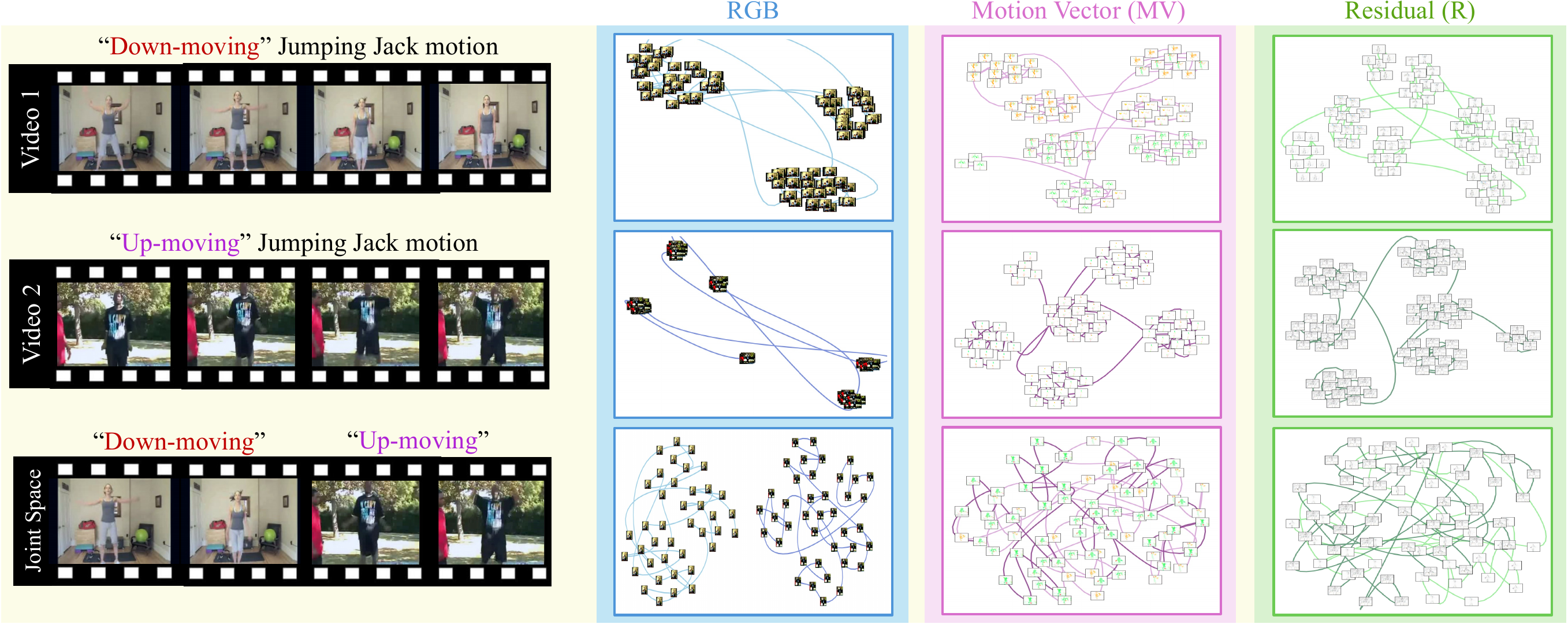}
    \vspace{-4pt}
    \captionof{figure}{We visualize two ``Jumping Jack" videos in t-SNE space~\cite{van2008visualizing} using their RGB and compressed domain (CD) (MV and R) representations to investigate learning benefits in CD space. Curves showing video trajectories. 
    While in joint RGB space the two videos are clearly separated, in MV and R space they overlap (row3). This suggests that a RGB-image based model needs to learn the two patterns separately, while a CD-based model sees a shared representation for videos of the same action, benefiting training and generalization. Also note that the two ways of the RGB trajectories overlap, showing that they cannot distinguish between the up- and down-moving motion. In contrast, CD signals preserve motion. The trajectories thus form circles instead of going back and forth on the same path (row1, row2).
    } 

    \label{fig:teaser}
\end{center} 
}] 
\normalsize
\begin{abstract}
Training robust deep video representations has proven to
be computationally challenging due to substantial decoding overheads, the enormous size of raw video streams, and their inherent high temporal redundancy. Different from existing schemes, operating exclusively in the compressed video domain and exploiting all freely available modalities, i.e., I-frames, and P-frames (motion vectors and residuals) offers a compute-efficient alternative. Existing methods approach this task as a naive multi-modality problem, ignoring the temporal correlation and implicit sparsity across P-frames for modeling stronger shared representations for videos of the same action, making training and generalization easier. By revisiting the high-level design of dominant video understanding backbones, we increase inference speed by a factor of $56$ while retaining similar performance. For this, we propose a hybrid end-to-end framework that factorizes learning across three key concepts to reduce inference cost by $330\times$ versus prior art: First, a specially designed dual-encoder scheme with efficient Spiking Temporal Modulators to minimize latency while retaining cross-domain feature aggregation. Second, a unified transformer model to capture inter-modal dependencies using global self-attention to enhance I-frame -- P-frame contextual interactions. Third, a Multi-Modal Mixer Block to model rich representations from the joint spatiotemporal token embeddings. Experiments show that our method results in a lightweight architecture achieving state-of-the-art video recognition performance on UCF-101, HMDB-51, K-400, K-600 and SS-v2 datasets with favorable costs ($0.73$J/V) and fast inference ($16$V/s). Our observations bring new insights into practical design choices for efficient next-generation spatiotemporal learners. Code is available.
\end{abstract}

\section{Introduction}
Videos command the lion’s share of internet traffic at
75\% and rising~\cite{pepper2013cisco}, posing an urgent need for efficient video understanding methods. A key task in video comprehension is to understand human actions, spanning across resource-constrained applications such as Virtual/Augmented Reality, autonomous navigation systems, and human-computer interaction~\cite{poquet2018video, lin2019tsm, geiger2012we}. 
However, present-day pipelines solving video action recognition fail to simultaneously reduce the computation burden while meeting performance requirements. We believe that the reason is two-fold. First, videos have
a very low information density -- $1$hr of a $720$p video can be compressed from $222$GB to $1$GB~\cite{wu2018compressed}. In other words, they are filled with repeating patterns, obscuring the `true' interesting signal. This redundancy makes it harder for methods to extract meaningful information, making training much slower. Second, with only RGB images, learning temporal structure is difficult. A vast
body of literature attempts to process videos as image sequences, with 2D-CNNs~\cite{karpathy2014large, yue2015beyond, simonyan2014two} and 3D-CNNs~\cite{carreira2017quo, feichtenhofer2016spatiotemporal, tran2015learning} on small-scale~\cite{soomro2012ucf101, kuehne2011hmdb} and largescale~\cite{kay2017kinetics, carreira2018short, goyal2017something, carreira2019short} benchmarks. Advancements in Vision Transformers~\cite{alexey2020image} offered long-range context modeling for image classification and video recognition~\cite{liu2022swin, liu2021swin, liu2022video, zhang2021vidtr, chen2024align, wang2023videomae, wasim2023video}. However, they are still bottlenecked by elevated energy consumption due to redundant information processing.


To address these issues, directly exploiting the compressed representation 
rather than operating on raw RGB frames offers an efficient alternative. Compression techniques (MPEG-4~\cite{duan2020omni}, H.264~\cite{richardson2011h} etc.) leverage temporal redundancy by recognizing that consecutive frames are often similar. They retain only a few frames completely (I-frames) and reconstruct other frames during decoding based on offsets from the complete images, called P-frames (motion vectors and residuals). This removes up to two orders of magnitude of superfluous information, making interesting signals prominent~\cite{le1991mpeg}. Also, the motion vectors (MV) in video compression implicitly encode the motion information absent in static RGB images, offering a richer representation of temporal dynamics. Moreover, P-frames are invariant to spatial variations, such as changes in clothing or lighting conditions, which ensures better generalization and the reduced variance further simplifies training(~\cref{fig:teaser}). 
Hence, learning directly from compressed videos can be much more efficient as we only look at the true signals in the sparse P-frames instead of repeatedly processing near-duplicates. Efficiency gains additionally come from bypassing video decoding: since videos are stored and transmitted in a compressed state, capitalizing on free MVs and residuals instead of raw frames avoids the decoding overhead.

\begin{figure}
\centering
\includegraphics[width=0.48\textwidth]{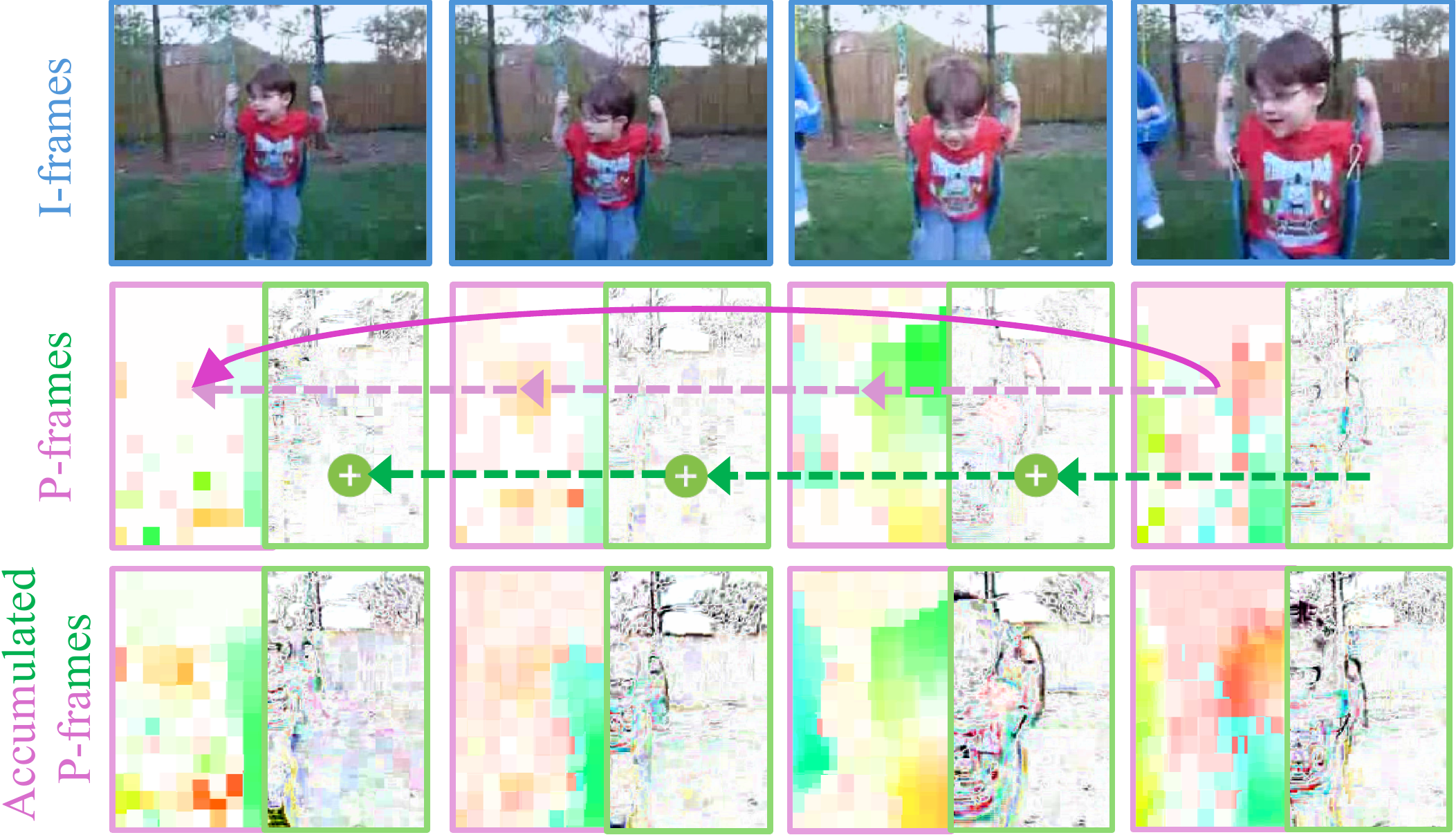} 
\vspace{-17pt}
\caption{Original motion vectors and residuals capture only inter-frame changes, often with low signal-to-noise ratios, making them difficult to model. In contrast, accumulated P-frames aggregate longer-term differences, revealing clearer temporal patterns.}
\vspace{-8pt}
\label{fig:acc}
\end{figure}

Existing methods trying to learn from compressed videos fall back on models originally designed for raw images~\cite{wu2018compressed, wang2021team, wu2020multi, dos2019cv}, and are inept at directly handling sparse and temporally correlated P-frame streams (~\cref{fig:acc}) since unlike RGB images, MVs and residuals are much sparser and lack dense semantic information. To further aggravate the issue, these methods discard the temporal ordering of inputs by representing
them as parallel channels and perform sub-optimal stateless computations on this sequential data stream~\cite{soufleri2024advancing, terao2023efficient}. Recent work has shown that such stateless processing shows inferior performance on compressed videos, improving only when it leverages handcrafted temporal mappings~\cite{liu2023learning}. Nevertheless, these methods require large model sizes and incur high inference costs ($>3.5J$ per video) since they do not rely on efficient processing to realize the sparse information retrieval aspect of P-frames.

With this in mind, we revamp the dominant architecture choices for video understanding by adapting them for compressed videos. To the best of our knowledge, this is the first work that jointly considers the temporal correlation and implicit sparsity within the compressed domain to reveal a set of key enablers for efficient video understanding at scale. To achieve this, we maximize accumulated \emph{analog} temporal context from MV and residual (R) streams using sparse implicit recurrence. In particular, we build a shallow Spiking Temporal Modulator (STM) module that does not rely on any of the earlier additional parametric and learning overheads. 
Next, different from naive spatial and temporal attention separation in the traditional transformer encoders~\cite{bertasius2021space, arnab2021vivit}, we propose jointly encoding context across space, time, and modality dimensions, to facilitate learning the inter-modal interactions. Further, we fuse the rich hierarchical spatio-temporal embeddings using our designed Multi-Modal Mixer to build a fast, lightweight and highly performant framework deployable on hybrid neuromorphic chips such as~\cite{pei2019towards}. Evaluations on five public action recognition datasets demonstrate that our method performs superior to the compressed domain counterparts, and comparably to established raw-domain methods with up to a dramatic $98\%$ energy savings and $55\%$ inference speed gains.


\section{Related Works}

\hspace{12pt}\textbf{Traditional Video Action Recognition} Recent video action recognition methods primarily operate on raw video data, feeding RGB frames directly into networks ~\cite{carreira2017quo, feichtenhofer2016convolutional, tran2015learning}. Earliest approaches relied on handcrafted features~\cite{klaser2008spatio, laptev2005space, wang2013dense}. However, with the breakthrough of 2D CNNs on ImageNet~\cite{krizhevsky2012imagenet, he2016deep, simonyan2014very, tan2019efficientnet}, these models were directly adapted for video recognition~\cite{karpathy2014large, yue2015beyond, simonyan2014two}. The release of large-scale datasets like Kinetics~\cite{kay2017kinetics} paved the way for 3D CNNs~\cite{carreira2017quo, feichtenhofer2016spatiotemporal, tran2015learning}, which significantly improved spatio-temporal modeling, surpassing 2D CNNs but at the cost of higher compute requirements. To mitigate this, various optimized 3D CNN architectures emerged~\cite{duan2020omni, feichtenhofer2020x3d, feichtenhofer2019slowfast, lin2019tsm, qiu2019learning, sun2015human, szegedy2016rethinking, xie2018rethinking}, improving efficiency and performance. 
Fully transformer-based architectures followed after~\cite{bertasius2021space, fan2021multiscale, li2022mvitv2, liu2022video, yan2022multiview}, achieving state-of-the-art performance across benchmarks. Recently, new hybrid CNN-ViT models~\cite{li2022uniformer} have been proposed, achieving performance on par with fully transformer-based methods.

\textbf{Compressed Video Action Recognition}
With a rising focus on exploiting information already available in compressed formats~\cite{smolic2000low} for video understanding, there have been recent efforts in this emerging research direction.

Some recent works look towards achieving high classification accuracy but at significant inference demands. The authors in~\cite{liu2023learning} propose cross-domain distillation to force the compressed-domain (CD) model to learn knowledge from the raw-domain (RD) model using traditional RNNs which are not designed for sparse and irregular P-frame data (also pointed out in~\cite{xie2018rethinking}). 
The authors of MFCD-Net~\cite{battash2020mimic} similarly propose mimicking the teacher network in the raw domain using a CD student network. For these works, the benefit of using compressed videos diminishes as they require access to the decoded raw video during training. 

An alternative thread of research restricts itself to leveraging only compressed videos for improved compute savings. CoViAR~\cite{wu2018compressed} was one of the earliest compressed video action recognition works in this category. CoViAR proposed training three 2D CNNs for each modality independently and naively averaged predictions from these three backbone networks at the inference phase. 
Follow-up work by TEAM-Net~\cite{wang2021team} uses a new module that is meant to capture the channel-wise and spatial relationship of the compressed modalities. These methods incurs significant parametric and compute overheads since they try to learn separate representations by ignoring the rich inter-modal relations, and end up discarding implicit sparsity and temporal ordering in the P-frames by processing them as parallel channels. As a result, they rely on multi-crop, multi-segment, and multi-clip evaluations for better performance, raising their inference latency and costs. Later CV-C3D~\cite{dos2019cv} extends 3D CNNs to the compressed domain using independent networks for the modalities. Improving upon their approach, DMC-Net~\cite{shou2019dmc} uses an additional network trained using Optical Flow (OF), which improves recognition performance at the increased cost of computing flow~\cite{kwon2020motionsqueeze, fan2019more, lin2019tsm}. Alternatively, authors in~\cite{wu2020multi, soufleri2024advancing} use the CoViAR backbone networks trained using multi-teacher distillation to compress their model. However, their distillation scheme across teachers is handcrafted, leading to a noticeable drop in accuracy although they achieve lower energy requirements. Inspired by the independent sub-network training in~\cite{havasi2020training}, ~\cite{terao2023efficient} proposes leveraging an efficient single network trained for multiple inputs-multiple-outputs (MIMO) by concatenating the MV, R, and I-frames as inputs. Their method reduces compute costs but imposes accuracy limitations constrained by the network’s capacity. Further, the coupled nature of the input forces the use of same frame count across all modalities, hindering real-time applications where the entire video stream may not be available in advance. Instead, MMViT~\cite{chen2022mm} proposes a naive ViT method to improve performance, incurring significant costs during inference. Our method builds on these preliminary approaches to instead propose a hybrid model unifying the advantages of different compressed signals in a compact way. We empirically demonstrate superior performance on five action recognition benchmarks, while being more computationally efficient and low latency. In addition, our examination provides key insights into the various proposed components of our method that enable these results.

\section{Method}

\subsection{Modeling Compressed Video Representations}
Common video compression algorithms enable highly efficient compression by splitting a video into I-frames (intra-coded
frames) and P-frames (predictive frames). An I-frame ($I$) is a regular image while a P-frame (MV and R) only holds changes in the image from the previous frame, thus saving space. A
MV, denoted by $\tau^t$, represents the movement of blocks of pixels (typically frames are divided into 16$\times$16 macroblocks during compression) from the source frame to the target frame at time $t$, thus roughly resembling coarse-grained optical flows. A residual $\delta^t$ encodes the difference between a P-frame and its reference I-frame after motion compensation from MVs, therefore showing changes in appearance and motion boundaries across time. Existing methods ignore the inherent sparsity and temporal dynamics in P-frames, processing them as regular frames.  Preliminary experiments suggest that their low signal-to-noise ratios also make them very difficult to model. These observations motivate us to investigate accumulated P-frames to aggregate longer-term differences, allowing better visualization of pixel ownership for moving objects (pixels on the same object move in the same direction, and generate spatially close iso-polarity vectors). Specifically, we trace all MVs back to the reference I-frame and accumulate the residuals (~\cref{fig:acc}). Given any pixel at location \( i \) in frame \( t \), let \(\mu_{\tau^{(t)}}(i) = i - \tau^{(t)}_i\)  be the referenced location in the previous frame. 
\begin{figure}[!t]
\centering
\includegraphics[width=0.48\textwidth]{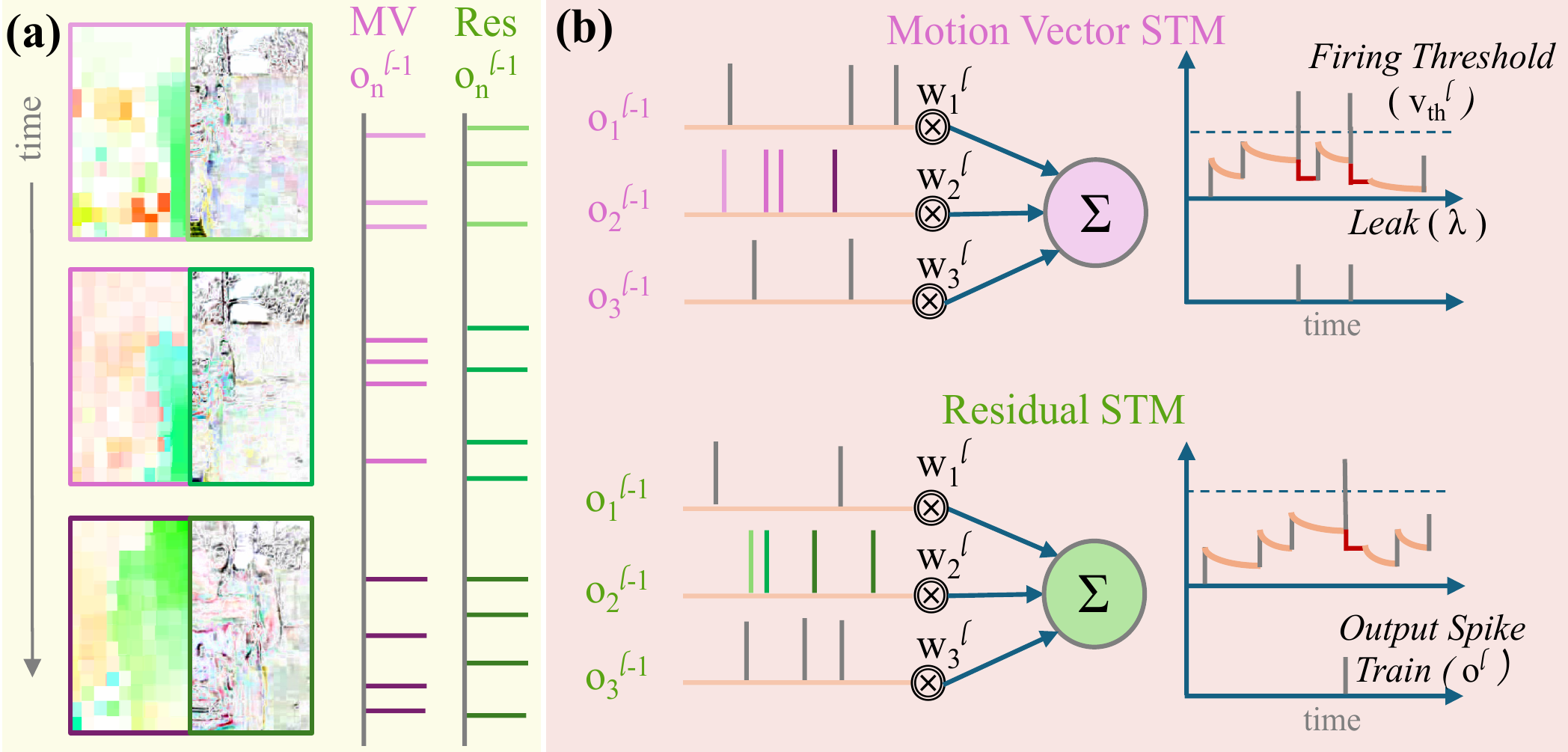} 
\vspace{-16pt}
\caption{Overview of the unrolled dynamics of our Spiking Temporal Modulator Unit. (a) MV and Residual inputs across time are (b) aggregated and modulated by synaptic weights to be integrated as current influx in the membrane potential for each STM. 
The firing threshold $v^l_{th}$ and the leak factor $\lambda^l$ are dynamically updated during training to attain best possible performance.}
\vspace{-11pt}
\label{fig:snn_feat}
\end{figure}
Our back-tracing technique expresses the location traced back to frame \( k < t \) as:
\small
\vspace{-4pt}
\begin{equation}
\Gamma_{i}^{(t,k)} = \mu_{\tau^{(k+1)}} \circ \mu_{\tau^{(k+2)}} \circ \dots \circ \mu_{\tau^{t}}(i)\\\label{eq1}
\vspace{-2pt}
\end{equation}
\normalsize
Accumulated MVs \(\Pi^{(t)} \in \mathbb{R}^{H \times W \times 2} \) 
and the accumulated residuals \(\Delta^{(t)} \in \mathbb{R}^{H \times W \times 3} \) at frame \( t \) can be defined as:
\small
\vspace{-6pt}
\begin{equation}
\Pi^{(t)}_i = i - \Gamma^{(t,k)}_i \text{  and  }
\Delta^{(t)}_i = \sum_{j=k+1}^{t-1} \delta^{(j)}_{\Gamma_i^{(t,j)}} + \delta^{(t)}_i\\ \label{eq2}
\vspace{-6pt}
\end{equation}
\normalsize
The original recurrence relation for reconstructing P-frames was \(I_i^{(t)} = I_{\mu_{\tau^{(t-1)}}(i)}^{(t-1)} + \delta_i^{(t)}\). Our proposed back-tracing refactors P-frames to now have a different dependency, which can be efficiently calculated in linear time through a simple feed forward algorithm as \(I_i^{(t)} = I_{i-\Pi_i^{(t)}}^{0} + \Delta_i^{(t)}\).
~\cref{fig:acc} shows the resulting accumulated P-frames. They exhibit less sporadic (noisy) and more expressive patterns than the original frames allowing for more pronounced temporal context to boost performance. We verify this in in~\cref{tmanalysis}.
\subsection{Spatio-Temporal Feature Modulation}
Recently, a new encoding method, Focal modulation~\cite{yang2022focal}
has been proposed. They follow an early aggregation process by First Aggregation, Last Interaction (FALI) mechanism, where context aggregation is performed first, followed by the interaction between the queries and the aggregated features. This in contrast to the First
Interaction, Last Aggregation (FILA) process used in state-of-the-art video
recognition methods~\cite{arnab2021vivit, bertasius2021space, fan2021multiscale, liu2022swin}. Refer to~\cite{yang2022focal} for an elaborate explanation. The focal modulation process given by~\cite{yang2022focal} works well on images by extracting the spatial context around a query token. However, to model spatio-temporal information, both the spatial and temporal context tokens have to be extracted. To achieve this, we propose our architecture which explicitly models
both I-frame \emph{(dense spatial)} and P-frame \emph{(sparse temporal)} information using a two-stream feature aggregation block in FALI style.

\textbf{Temporal Modulator} To leverage the sparsity of P-frames, we propose a Spike-based Temporal Modulator (STM) with Conv-Leaky-Integrate-and-Fire (ConvLIF)~\cite{abbott1999lapicque, dayan2001theoretical} layers that use a biologically-inspired mechanism to retain and `spike’ relevant features over time, filtering unnecessary temporal noise and skipping computation on neuromorphic hardware in the absence of inputs.
We devise the dynamics of the neuron model as follows:
\small
\begin{equation}
    u_{mem}^{l}[t] = w^{l}o^{l-1}[t] +\lambda^{l}u_{mem}^{l}[t-1]-v_{th}^{l}o^{l}[t-1]\\
    \label{eq1}
\end{equation}
\begin{equation}
o[t] = \mathcal{H}(u_{mem}[t] - v_{th})
\end{equation}
\normalsize
where $\mathcal{H}$ represents the Heaviside step function~\cite{weisstein2002heaviside}. At timestep $t$, weighted output spikes from the previous neuron $l-1$ are accumulated in the membrane potential $u_{mem}^l[t]$ of the neuron $l$ creating a `short-term memory' (~\cref{fig:snn_feat}). At the same time, $u_{mem}^l[t]$  of the neuron $l$ decays by a leak factor $\lambda^l$  to represent `forgetting'. Once the accumulated membrane potential exceeds the firing threshold $v^{l}_{th}$, the neuron generates a binary spike output ($o^l$). $u_{mem}^l[t]$ is reset after all the P-frames are processed. We regard this sparse potential accumulation, decay, and resting process as an efficient \textit{temporal modulator}. However, for the aggregated temporal embeddings to have higher expressibility than prior attempts~\cite{lee2020spike} that use binary spike trains  to transmit temporal features, we propose to use the analog membrane potential to transmit neural activations from the STM, while still demanding drastically lower compute compared to traditional RNNs. This allows our model to avoid performance degradation from over-sparsification of the feature maps while still retaining all benefits of spike-based processing.
 
Dynamic threshold schemes observed in different biological systems provide essential inspiration for increasing robustness to noise or camera motion~\cite{fontaine2014spike, pozo2010unraveling}. To address low signal-to-noise ratios observed in P-frames, we propose to learn the $v_{th}$ and $\lambda$ for each layer dynamically to find the optimal hyperparameters. Allowing our network to learn on the go how `important' a new input is, or how much of the past motion influences its current state by adjusting its threshold and leak enables our dynamical STM to robustly model both short- and long-range dependencies within a video with improved compute and parameter efficiency. By decoupling the temporal branches for MVs and residuals, we are able to separately extract and aggregate temporal context for each, leading to boosts in performance. We further examine these aspects in~\cref{tmanalysis} and the appendix.

\textbf{Spatial Modulator} The local-spatial feature extraction stage should incorporate a prior that pixels are arranged in a 2D grid as early as possible in the computation graph. Here, we propose to aggregate the neighboring spatial context for each visual query by decomposing I-frames and temporal modulated P-frames into $N$ non-overlapping patches using learnable convolution on the features instead of the standard linear embedding layers as used in ViTs~\cite{dosovitskiy2020image}. Hence, unlike prior work, our resulting token sequences can enjoy added flexibility by no longer being tied to an input resolution strictly divisible by a pre-set patch size.

\begin{figure*}[!th]
\centering
\includegraphics[width=0.94\textwidth]{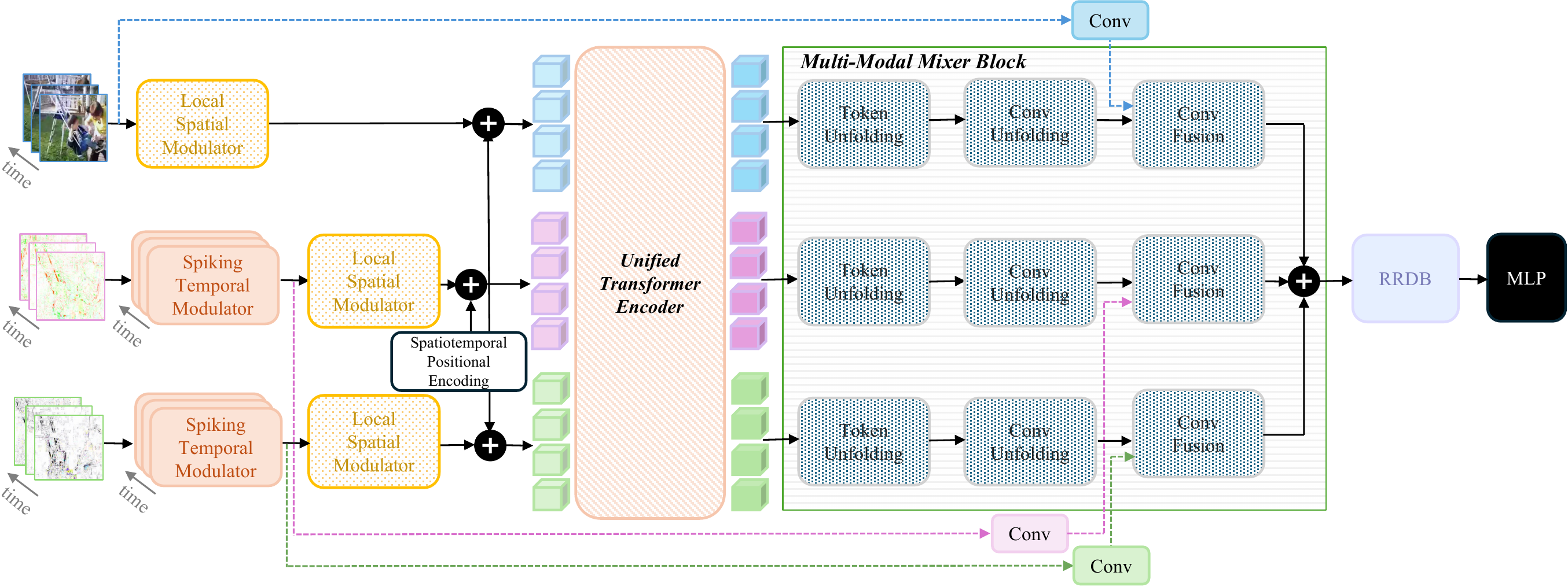} 
\vspace{-8pt}
\caption{Overview of our method for unified representation learning in compressed videos. Given a set of compressed inputs, our factorized encoder modules sequentially aggregate rich spatio-temporal embeddings. These then interact to capture both local and global dependencies across the modalities and are fused using a specially designed Multi-Modal Mixer Block at different levels of granularity.
}
\vspace{-5pt}
\label{fig:arch}
\end{figure*}

\subsection{Cross Space-Time-Modality Contextualization}
As discussed above, our approach aims to solve both large local redundancies and complex global dependencies for effective and efficient video understanding by learning sequentially along space-time dimensions across the modalities. 
First, gathered temporal and local-spatial contexts from the Temporal and Spatial Modulators are fused to build the final feature tokens such that our design adapts the desirable qualities of early aggregation in the FALI style to video tasks. In the subsequent step, we add Spatiotemporal Positional Encodings~\cite{chen2022mm} to maximize context on each token and perform dilated global feature mixing using self-attention in the Unified transformer to aggregate information from coarser dependencies within the concatenated inputs. This decoupled factorization across time, space, and modality dimensions hierarchically for the modalities allows for modeling both short- and long-range dependencies in a video with low computational and parameter overheads. This is followed by reshaping the visual tokens back to their original dimensions using Token Folding technique to ensure the correct arrangement of all patches. These folded tokens are passed through a Multi-Modal Mixer Block, which employs overlapping convolution operations in the Conv. Folding modules and skip connections for information interaction across multiple levels
of granularity. For this, we fuse aggregated high-level spatio-temporal maps from the Conv. Folding stage with prior low-level temporal feature maps after channel-mixing (pointwise convolutions) in the Conv. Fusion block to reinforce strong temporal context. 

The condensed cross-modal features are then directed into a residual in residual dense block (RRDB)~\cite{wang2018esrgan} which introduces residual connections within and between multiple dense blocks, increasing the network’s depth and complexity. Global average-pooling (GAP) of all the tokens followed by a linear classifier $\in \mathbb{R}^{d\times C}$ is used to classify the final output representation to one of $C$ classes instead of a class token~\cite{beyer2022better, alexey2020image} since we find that global feature aggregation benefits tasks like classification.~\cref{fig:arch} shows the overall method and we validate its effectiveness via detailed ablations and comparisons to alternate designs in~\cref{tmanalysis}.

\begin{table*}[!h]
\caption{Comparison on UCF-101 and HMDB-51 to state-of-the-art methods 1) accessing decoded videos (yellow), 2) requiring pre-computed optical flow (purple) 3) using only compressed videos (orange). CD: Compressed Domain, RD: Raw Domain, OF: Optical Flow. For “views”, $x\times y$ denotes $x$ temporal views and $y$ spatial crops. Parameter efficiency and inference energy cost are computed across all views on the standard 45nm CMOS process~\cite{horowitz20141} (See appendix for details). We use an RTX $2080$ Ti GPU for speed comparisons.}
\vspace{-5mm}
\begin{center}
\resizebox{1.0\textwidth}{!}{
\begin{tabular}{lcc|cc|ccccc}
\hline \hline
 Methods & Modality  & Views & UCF-101 [\%] &  HMDB-51 [\%]& $\text{Params} (M)$ & $\text{TFLOPs}_\text{ANN}$ & $\text{GFLOPs}_\text{Spike}$ & $\text{E}_\text{Total} (J)$ &  $\text{Speed (V/s)}$\\ \hline \vspace{-4mm} \\
\rowcolor{LighterPastelYellow}
 ALT-L/14~\cite{chen2024align}& RD   & 32$\times$1 &98.1 & 79.1& 437.1& 52.77 & - & 242.733& -\\ 
 \rowcolor{LighterPastelYellow}
 VidTr~\cite{zhang2021vidtr}& RD   & 10$\times$3 & 96.7 & 74.4 & 90.89 & 11.13 & - & 51.180& 0.284\\ \hline
 \rowcolor{LighterPastelYellow}
 Liu et. al~\cite{liu2023learning} & RD + CD  &  8$\times$10 & 95.8& 73.5& $>$81.72 & $>$0.74& - &$>$3.422&-\\
 \rowcolor{LighterPastelYellow}
 MFCD-Net~\cite{battash2020mimic} & RD + CD  & 12$\times$15 & 93.2& 66.9&16 & 0.13& - & 0.589& 0.352\\ \hline
\rowcolor{LighterPastelPurple}
 DMC-Net~\cite{shou2019dmc} &  CD + OF & 8$\times$10 & 92.3& 71.8 &92.4 & 2.99 & - & 13.771& 0.043\\ \hline
\rowcolor{LightOrange}
CV-C3D~\cite{dos2019cv} & CD  &16$\times$1  & 83.9&55.7 &235.3 & 0.11 & - & 0.501& 1.259\\
\rowcolor{LightOrange}
 MIMO~\cite{terao2023efficient} & CD  &  8$\times$1& 89.2& 62.9&37.09 & 0.81& - & 3.709& 1.307\\
\rowcolor{LightOrange}
 TEAM-Net~\cite{wang2021team} & CD  & 8$\times$10& 94.3&73.8 &48.65 & 0.81& - & 3.714& 1.537\\
\rowcolor{LightOrange}
CoViAR~\cite{wu2018compressed} & CD  & 8$\times$10 & 90.4& 59.1&81.72 & 0.74& - & 3.422& 1.585\\
\rowcolor{LightOrange}
Wu et al.~\cite{wu2020multi} & CD  & 8$\times$10 & 88.5&56.2 &33.69 & 0.45& - & 2.075& 1.658\\
\rowcolor{LightOrange}
PKD~\cite{soufleri2024advancing} &  CD & 3$\times$1 & 88.4&60.3 &79.2 & 0.12 & - & 0.544& -\\
\rowcolor{LightOrange}
MM-ViT~\cite{chen2022mm} &  CD &  8$\times$3& 93.3& -&- &  18& - & 82.801 & -\\
\rowcolor{LightOrange}

\emph{\textbf{Ours}} & \textbf{CD}  & \textbf{8$\times$1} & \textbf{95.6}& \textbf{74.6}&\textbf{83.41} & \textbf{0.15} & \textbf{3.99} & \textbf{0.734} &\textbf{16.025}\\ \hline \hline \vspace{-8mm} 
\end{tabular}}
\end{center}
\vspace{-2mm}
\label{ucftable}
\end{table*}

\section{Experimental Evaluation}
\subsection{Setup and Protocols}

\hspace{12pt}\textbf{Datasets} We evaluate on five popular video benchmarks: UCF-101~\cite{soomro2012ucf101}, HMDB-51~\cite{kuehne2011hmdb},
Kinetics-400~\cite{kay2017kinetics}, Kinetics-600~\cite{carreira2018short} and Something-Something-V2~\cite{goyal2017something}. UCF-101 contains $13,320$ trimmed videos from $101$ action categories while HMDB-51 contains $6,766$ videos from $51$ categories. Each dataset has three training-testing splits. We report the average top-1 performance across the three splits. K-400 and K-600 are large-scale YouTube video datasets, having $400$ and $600$ classes respectively, while SS-V2 consists of about $220$K videos with a duration of $2$ to $6$ seconds for $174$ categories. 
Following~\cite{wang2021team}, we use MPEG-4 encoded videos where an I-frame is followed by $11$ P-frames.

\textbf{Implementation Details.} Following TSN~\cite{wang2016temporal}, we resize all training videos to $340\times256$. Then, random horizontal flipping and scale-jittering were utilized for data augmentation. Each cropped frame was finally resized to $224\times224$ for training the network. 
Patch size was set to $16\times16$ across all visual modalities. For UCF-101, HMDB-51, K-400 and K-600 datasets, we follow a similar training scheme to~\cite{wang2021team} and train for $300$ epochs using Adam optimizer~\cite{diederik2014adam} with a weight decay of $0.0001$, an epsilon value of $0.001$ and an initial learning rate of $0.02$. For the SS-v2, K-400 pretrained weights were used to initialize the model. 
However, unlike standard backpropagation~\cite{rojas1996backpropagation}, gradient computation in the proposed STM is not straightforward since LIF neurons have a spiking mechanism that generates
non-differentiable threshold functions. We enable learning in this module with surrogate gradients to approximate the gradient of the Heaviside step function during backpropagation~\cite{neftci2019surrogate, lee2016training} and use an inverse tangent surrogate gradient function with width $\gamma = 100$ (to allow sufficient gradient flow) since it is computationally inexpensive. During inference, we report results as an average across $8$ uniformly sampled triplets of I-frames, motion vectors, and residuals with a center-crop strategy. To fairly estimate energy costs, we use the number of floating point operations (FLOPs) performed by the network across total clips sampled from the video and spatial crops taken per video during inference. \emph{Obvious energy and time savings from avoiding video decompression are not included and come in addition to the reported costs}. More details are available in the appendix.




\begin{table}[t]
\caption{Comparison on SS-v2, K-400 and K-600 datasets.}
\vspace{-6mm}
\begin{center}
\resizebox{1.0\linewidth}{!}{
\begin{tabular}{lc|ccc|c}
\hline \hline
Method & Modality & SS-v2 [\%] & K-400 [\%]  & K-600 [\%]&$\text{E}_\text{Total} (J)$ \\ \hline \vspace{-4mm} \\
\rowcolor{LighterPastelYellow}
 ViViT-L FE{~\cite{arnab2021vivit}} &RD&  65.9& 81.7& 83.0&54.92 \\
 \rowcolor{LighterPastelYellow}
 VidTr~\cite{zhang2021vidtr} & RD& 63.0 & 77.7  & -& 51.18\\ 
\rowcolor{LighterPastelYellow}
I3D~\cite{carreira2017quo} & RD&  50.0 & - & &100.67\\ 
\rowcolor{LighterPastelYellow}
VideoMAE V2-g~\cite{wang2023videomae} & RD& 77.0&88.5 & 88.8 & 2633.04\\
\rowcolor{LighterPastelYellow}
VideoSwin-L~\cite{liu2022video} & RD&69.6 & 84.9 & 86.1 & 484.61\\
\rowcolor{LighterPastelYellow}
Video-FocalNet-B~\cite{wasim2023video} & RD& 71.1 & 83.6 & 86.7 & 822.48\\ \hline
\rowcolor{LighterPastelYellow}
Liu et. al~\cite{liu2023learning} &RD + CD& - & 73.5 &- &$>$3.42 \\
\rowcolor{LighterPastelYellow}
MFCD-Net~\cite{battash2020mimic}&RD + CD & - & 68.3 & -&0.59\\ \hline
\rowcolor{LightOrange}
CoViAR~\cite{wu2018compressed} &CD& - & 69.1&- & 3.42 \\ 
\rowcolor{LightOrange}
TEAM-Net~\cite{wang2021team} &CD& - &72.2 & -& 3.71 \\ 
\rowcolor{LightOrange}
MM-ViT~\cite{chen2022mm} &CD& 64.9 &- & 81.5& 82.80 \\ 
\rowcolor{LightOrange}
\emph{\textbf{Ours}}& \textbf{CD} & \textbf{62.5}  & \textbf{74.2}  & \textbf{80.3} & \textbf{0.73} \\
\rowcolor{LightOrange}
\emph{\textbf{Ours-L}}& \textbf{CD} &  \textbf{64.2} & \textbf{75.9}  & \textbf{81.1} & \textbf{1.40}\\ \hline \hline \vspace{-10mm} 
\end{tabular}}
\end{center}
\vspace{-5pt}
\label{k400table}
\end{table}

\subsection{Benchmark Comparisons}

\subsubsection{Quantitative Results}

\begin{figure*}[!t]
\centering
\includegraphics[width=0.96\textwidth, height=0.35\textheight]{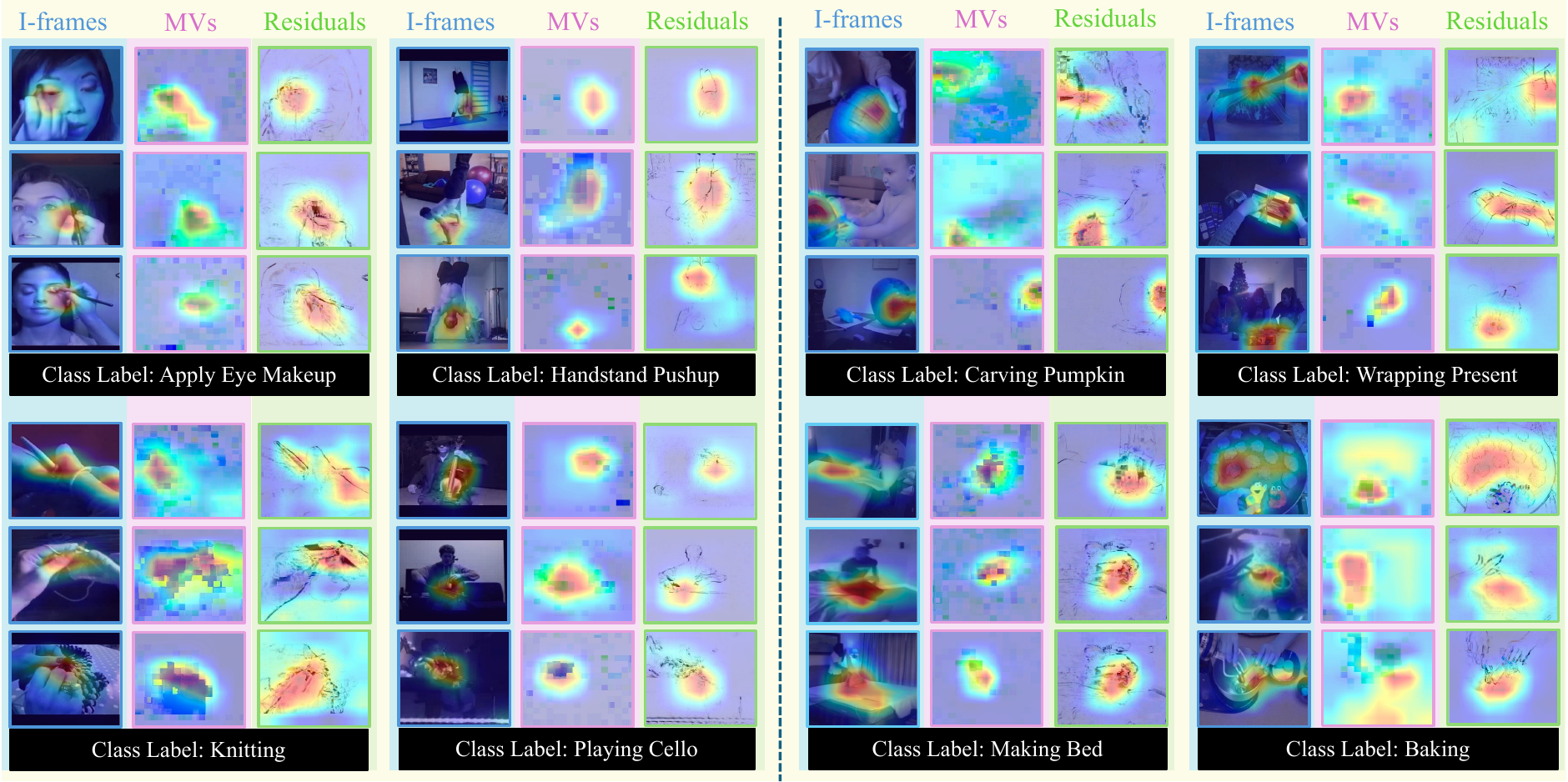} 
\vspace{-8pt}
\caption{Visualization of significant features extracted by our method to the input space across multiple action classes for the UCF-101 dataset (left) and K400 dataset (right). Our proposed model learns to focus on relevant parts of the video for classification across modalities.}
\vspace{-5pt}
\label{fig:heat_map}
\end{figure*}

\begin{figure}
\centering
\includegraphics[width=0.43\textwidth, height=0.20\textheight]{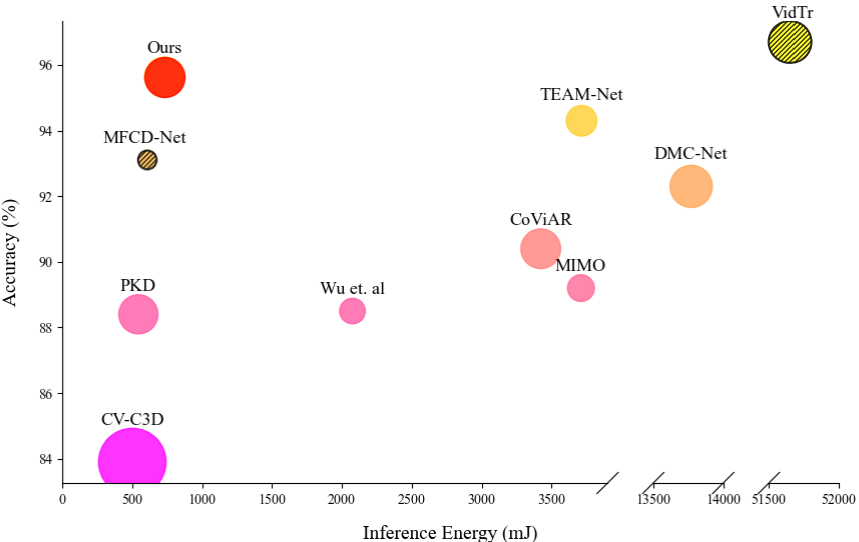} 
\vspace{-8pt}
\caption{Performance vs Inference energy on UCF-101 evaluated on the 45nm CMOS process. Circle areas are proportional to the model size. Hatched circles denote models using RD videos.}
\vspace{-9pt}
\label{fig:plot_comp}
\end{figure}

\hspace{6pt}\textbf{UCF-101} ~\cref{ucftable} presents detailed results on UCF-101~\cite{soomro2012ucf101}. We compare to existing methods in CD~\cite{wang2021team, wu2018compressed, wu2020multi, terao2023efficient, soufleri2024advancing, dos2019cv, shou2019dmc} and find that our framework leverages complementary \emph{dense-spatial sparse-temporal} learning effectively to consistently outperform them by up to $14\%$ and achieves new state-of-the-art performance at $19\times$ lower inference costs. This accuracy is also higher or comparable to established RD alternatives that use the decoded single RGB modality~\cite{chen2024align, zhang2021vidtr, liu2023learning, battash2020mimic}. This confirms that the additional MV and R inputs provide important complementary motion features, which benefit learning while circumventing traditional decoding overheads. We qualitatively visualize these benefits in~\cref{fig:heat_map}. Our model is the smallest among literature by a large margin and still achieves better accuracy than~\cite{chen2022mm} while using lesser views and compute, making it a top bidder for energy-efficient applications, as visualized in~\cref{fig:plot_comp}. Finally, we evaluate network inference speeds across methods and find that our lightweight approach offers significantly lower latency than all counterparts.

\textbf{HMDB-51} We summarize evaluation results for HMDB-51~\cite{kuehne2011hmdb} in~\cref{ucftable}. Our method outperforms existing methods that use CD videos with or without optical flow by $0.05\%$ and $0.02\%$, respectively, while using up to a significant $95\%$ lower energy and $372\times$ higher inference speed. This validates that our highly lightweight framework can efficiently leverage spatiotemporal context from all modalities to boost performance. Remarkably, we also surpass established RD methods~\cite{zhang2021vidtr, liu2023learning, battash2020mimic} that cannot process videos efficiently in the compressed domain while requiring up to $70\times$ lower compute, besides added savings in decoding time and costs.

\textbf{K-400, K-600} The performance of our method on the Kinetics-400 and 600 datasets~\cite{kay2017kinetics, carreira2018short} is reported in~\cref{k400table}. Since they are more challenging datasets, we also train a bigger variant of our model `Ours-L' in addition to the base model `Ours'. Parameter details for the models are reported in the appendix. Compared to the prior state-of-the-art TEAM-Net~\cite{wang2021team} in the compressed domain, our method improves performance at $62\%$ lower costs. This is primarily because they do not leverage a temporal module for capturing the sparse yet rich motion information in P-frames, treating them as stateless inputs. Popular RD methods~\cite{zhang2021vidtr, arnab2021vivit, wang2023videomae, liu2022video} as well as methods that need access to raw videos during training~\cite{liu2023learning, battash2020mimic} claim comparable results, albeit at the cost of $97\%$ higher inference energy per video. This can be mainly attributed to the inherent self-recurrence in our STMs, which are more suitable to denoise and extract sparse cues compared to regular CNNs/RNNs/ViTs not designed for the sparse or irregular temporal data present in compressed formats~\cite{xie2018rethinking}.

\textbf{SS-v2} We report results on the SS-v2~\cite{goyal2017something} benchmark in~\cref{k400table} and note that amongst methods not relying on RD to boost performance, we perform competitively to~\cite{chen2022mm} while using a significant $99\%$ lower cost. Our method also surpasses the previous state-of-the-art in the raw domain~\cite{carreira2017quo} on this temporally challenging benchmark by up to $0.3\%$. Contemporaries such as~\cite{arnab2021vivit, zhang2021vidtr, wang2023videomae, liu2022video} similarly using decoded videos claim a slightly higher Top-1 accuracy, while using a much larger network and exorbitant inference costs ($\sim 73.5\%$ higher) in addition to their overhead from decoding videos to even begin processing. This strong performance shows that our method can effectively model the subtle temporal changes and dependencies in this challenging dataset while being highly lightweight, making it an obvious choice for energy-efficient edge applications.

\subsubsection{Qualitative Results}
To qualitatively evaluate our approach, we examine feature maps extracted by our method. In~\cref{fig:heat_map}, we show examples obtained from UCF-101 and K-400 datasets. We see that our method indeed attends to the relevant regions in the input space. For example, when applied to classify the video for `Knitting', the model concentrates on the fingers wielding the needle and the needle itself. In some cases, we find that our method benefits from representations in all the $3$ modalities; while classifying `Wrapping Present' (row $1$), the model focuses on the wrapping bow in the I-frame, and either hand in the temporally rich P-frames to learn a more holistic representation of the whole action instead of just relying on the presence of a ribbon/bow to make predictions for this class.
This strong alignment between quantitative and qualitative results reaffirms the effectiveness of the proposed method in complex spatial-temporal reasoning.

\textbf{Generalization using compressed signals} In~\cref{fig:teaser}, we qualitatively study the feature representations of two videos (up/down moving motions) of the same action in t-SNE space~\cite{van2008visualizing} to investigate feature generalization across an action. In joint video space for RGB images, we find that the two video trajectories are separated, while they overlap for P-frame images. This suggests that with compressed signals, videos of the same action can share statistical strength better irrespective of motion order. Also, the RGB trajectories keep retracing the same path, indicating that they fail to differentiate between the opposite motion directions in the videos. Conversely, CD signals retain motion information, creating repeating circular patterns. This helps boost our performance by modeling richer shared context across different videos for a given action, benefiting generalization.




\section{Ablations}

\hspace{12pt}\textbf{Effect of input modality} To evaluate the importance of each data modality, we perform an ablation study by training with different modality combinations on UCF-101, as shown in~\cref{modalitytable}. Unsurprisingly, we find that MVs and residual frames are essential for video recognition. Excluding either modality leads to an accuracy drop up to $3\%$ since they provide orthogonal information that complements the spatial context of static images. Still, I-frames are the most impactful data modality due to their dense spatial context; removing them alone decreases top-1 accuracy by $12\%$.

\textbf{Effect of Contextualization Order} We examine the effect of context learning sequence by enumerating various possible orders of the local spatial (learnable patch embedding), temporal (STM), and global spatial attention (unified transformer encoder) on UCF-101. As seen in~\cref{stmtable_b}, 
applying temporal feature extraction (T) before local spatial (LS) and global spatial (GS) attention yields slightly but consistently better performance than the reverse. This can be attributed to the benefit of temporal representations from P-frames, which provide early cues for distinguishing between actions that share similar appearance features.

\textbf{Effect of Unified Learning} To understand the advantages of unified global-spatial attention across modalities, we conduct a series of ablation experiments on the UCF-101 dataset using isolated transformer encoders for each input. In the first setting, we introduce cross-attention between \emph{separate} uni-modal transformer encoders where queries can only interact with keys and values from other modalities. The next setting uses \emph{separate} transformer encoders for each modality, followed by the fusion of their
respective outputs. The last setting is our proposed method with a \emph{unified} encoder which takes the concatenation of tokens across modalities to jointly develop a rich spatio-temporal global context.~\cref{stmtable_a} summarizes performance across these variants at their required compute costs. We find that our unified model achieves the best accuracy at minimal overhead by effectively capturing inter-modal dependencies.

\textbf{Temporal Modulator Analysis}\label{tmanalysis} In~\cref{stmtable_c}, we underscore the importance of effectively learning sparse temporal representation in the P-frame modalities. Removing the STM, using a common STM for both P-frame inputs, or not making $v_{th}$ and $\lambda$ dynamic degrades performance. When replacing the Spiking layers in the STM with an LSTM~\cite{shi2015convolutional} with one conv. layer per cell, we still achieve better performance. This suggests that though LSTMs can also extract temporal cues, the spiking mechanism of the STM acts not only as temporal memory but also as a natural filter to noise or camera motion, which is beneficial for robust predictions.

\begin{table}[!t]
\caption{Impact of different modality
combinations on UCF-101.}
\vspace{-5mm}
\begin{center}
\resizebox{0.3\textwidth}{!}{
\begin{tabular}{lcc|cc}
\hline \hline
 I & MV & R & Top-1 [\%] & Top-5 [\%]\\ \hline  
\ding{51} & \ding{51} & \ding{55} & 93.21 & 97.46  \\
\ding{51} & \ding{55} & \ding{51} & 92.93 &  97.01 \\
\ding{55} & \ding{51} & \ding{51} & 84.06 & 88.32  \\

\rowcolor{LightOrange}
\hline
\ding{51} & \ding{51}& \ding{51} & 95.63 & 99.53\\ \hline \hline \vspace{-8mm} 
\end{tabular}}
\end{center}
\label{modalitytable}
\end{table}

\begin{table}
\caption{Ablations to evaluate our proposed key components.}
\vspace{-2mm}
    \centering
    \resizebox{0.5\textwidth}{!}{
        \begin{minipage}{0.29\textwidth}
            \centering
            \begin{tabular}{l|c}
                \hline \hline
                Variant & Top-1 [\%] \\ \hline 
                LS $\Rightarrow$ GS $\Rightarrow$ T  & 90.72  \\
                LS $\Rightarrow$ T $\Rightarrow$ GS &  91.64 \\
                \rowcolor{LightOrange}
                \hline
                T $\Rightarrow$ LS $\Rightarrow$ GS  &  95.63 \\
                \hline \hline
            \end{tabular}
            \subcaption{Contextualization Order}
            \label{stmtable_b}
        \end{minipage}%
        \hfill
        \begin{minipage}{0.31\textwidth}
            \centering
            \begin{tabular}{l|cc}
                \hline \hline
                Variant & Top-1 [\%] & $\text{E}_\text{Total} (J)$ \\ \hline 
                Cross-att. & 94.72 & 0.93\\
                Separate & 94.18  & 0.95\\
                \rowcolor{LightOrange}
                \hline
                Unified  & 95.63& 0.73\\
                \hline \hline
            \end{tabular}
            \normalsize
            \subcaption{Transformer Encoder Analysis}
            \label{stmtable_a}
        \end{minipage}
    }
    
    \vspace{3pt}
    
    \resizebox{0.5\textwidth}{!}{
        \begin{minipage}{0.32\textwidth}
            \centering
            \begin{tabular}{l|c}
                \hline \hline
                Variant & Top-1 [\%] \\ \hline 
                W/o STM &  88.61 \\
                Common STM & 93.45\\
                STM $\rightarrow$ LSTM & 93.92  \\
                STM (fixed $v_{th}$, $\lambda$) &  93.91 \\
                \rowcolor{LightOrange}
                \hline
                STM  &  95.63\\
                \hline \hline
            \end{tabular}
            \subcaption{Temporal Modulator Analysis}
            \label{stmtable_c}
        \end{minipage}%
        \hfill
        \begin{minipage}{0.32\textwidth}
            \centering
            \begin{tabular}{l|c}
                \hline \hline
                Variant & Top-1 [\%] \\ \hline 
                W/o P-frame Accumulation & 89.65  \\
                W/o Multi-Modal Mixer  & 90.84 \\
                W/o skips & 89.02  \\
                W/o RRDB &  91.77 \\
                GAP $\rightarrow$ Class Token & 94.31  \\
                \hline \hline
            \end{tabular}
            \subcaption{Model Variants}
            \label{stmtable_d}
        \end{minipage}
    }
    \vspace{-5mm}
    \label{ablation_studies}
\end{table}

\textbf{Inspecting model variants} Finally, as shown in~\cref{stmtable_d}, we examine the key components and design choices of our proposed method. We find that removing the skip connections, RRDB module, Multi-Modal Mixer block, or using a classification token instead of Global Average Pooling (GAP) does not work well for the task. Also, we validate the benefits of our P-frame accumulation technique that not only simplifies the temporal dependencies but also creates clearer patterns to model, leading to boosts in performance.


\section{Conclusion}

To efficiently learn robust spatio-temporal representations that can effectively model both local and global contexts, this paper re-examines the design of established video understanding backbones. Motivated by the practical observation that decompressing videos is not only an overhead but also an inconvenience since it makes representations less robust and increases the dimensionality to make training computationally challenging, we propose a lightweight yet powerful factorized end-to-end framework to unify the advantages of compressed video modalities in a compact way. The efficacy of our design choices in effectively modeling shared spatiotemporal statistical patterns in the compressed representation is evidenced by our strong performance on five public benchmarks while offering sizeable benefits in computational cost and inference latency. The resulting design is an excellent choice for resource-constrained edge applications and \emph{hopes to inspire future work toward efficient video understanding systems not requiring decoded videos}.

{\small
\bibliographystyle{ieeetr}
\bibliography{main}
}

\end{document}